\documentclass[lettersize,journal]{IEEEtran}
\usepackage{amsmath,amsfonts}
\usepackage{algorithmic}
\usepackage{algorithm}
\usepackage{array}
\usepackage[caption=false,font=normalsize,labelfont=sf,textfont=sf]{subfig}
\usepackage{textcomp}
\usepackage{stfloats}
\usepackage{url}
\usepackage{verbatim}
\usepackage{graphicx}
\usepackage{cite}
\usepackage{multirow}
\usepackage{booktabs}
\usepackage{bm}
\usepackage{color}
\usepackage{makecell}
\begin{document}

\title{Sky-GVIO: an enhanced GNSS/INS/Vision navigation with FCN-based sky-segmentation in urban canyon}

\author{Jingrong Wang, Bo Xu, Ronghe Jin, Shoujian Zhang, Xingxing Li, Kefu Gao, and Jingnan Liu
\thanks{
This work was supported by the National Key Research and Development Program of China under Grant 2021YFB2501100. (Corresponding author: Shoujian Zhang.)
}
\thanks{Jingrong Wang, Kefu Gao and Jingnan Liu are with the GNSS research center, Wuhan University, Wuhan 430079, China;}
\thanks{Bo Xu, Shoujian Zhang and Xingxing Li are with School of Geodesy and Geomatics, Wuhan University, Wuhan 430079, China; (email: shjzhang@sgg.whu.edu.cn).}
\thanks{Ronghe Jin is with The Department of Aeronautical and Aviation Engineering, Hong Kong Polytechnic University, Hong Kong, China}}



\maketitle

\begin{abstract}
Accurate, continuous, and reliable positioning is a critical component of achieving autonomous driving. However, in complex urban canyon environments, the vulnerability of a stand-alone sensor and non-line-of-sight (NLOS) caused by high buildings, trees, and elevated structures seriously affect positioning results. To address these challenges, a sky-view images segmentation algorithm based on Fully Convolutional Network (FCN) is proposed for GNSS NLOS detection.  Building upon this, a novel NLOS detection and mitigation algorithm (named S-NDM) is extended to the tightly coupled Global Navigation Satellite Systems (GNSS), Inertial Measurement Units (IMU), and visual feature system which is called Sky-GVIO, with the aim of achieving continuous and accurate positioning in urban canyon environments. Furthermore, the system harmonizes Single Point Positioning (SPP) with Real-Time Kinematic (RTK) methodologies to bolster its operational versatility and resilience. In urban canyon environments, the positioning performance of S-NDM algorithm proposed in this paper is evaluated under different tightly coupled SPP-related and RTK-related models. The results exhibit that Sky-GVIO system achieves meter-level accuracy under SPP mode and sub-decimeter precision with RTK, surpassing the performance of GNSS/INS/Vision frameworks devoid of S-NDM. Additionally, the sky-view image dataset, inclusive of training and evaluation subsets, has been made publicly accessible for scholarly exploration at https://github.com/whuwangjr/sky-view-images .  
\end{abstract}

\def\abstractname{Note to Practitioners}
\begin{abstract}
This study focus on the tight integration of multiple homogeneous and heterogeneous sensors (e.g. GNSS/INS/Vision) with the goal of addressing GNSS NLOS interference challenges for wide-area vehicle navigation applications in urban canyon. We propose a sky-view images-based accurate and efficient NLOS detection and mitigation algorithm (named S-NDM), and extend it to the tightly coupled GNSS/INS/Vision integration framework (called Sky-GVIO). The LOS/NLOS satellites are identified by associating the semantic information of sky-view images, and a reasonable stochastic model is constructed to suppress NLOS influence in the tightly coupled GNSS/INS/Vision integration model positioning accuracy. The experimental results explains that the Sky-GVIO is able to maximize the use of as much sensor information as possible to achieve accurate and robust positioning in the real urban canyon scenarios.
\end{abstract}


\section{Introduction}
\IEEEPARstart{A}{utonomous} driving is one of the significant components in the field of intelligent transportation, necessitating high-precision localization, notably within the challenging terrains like urban canyons. Currently, the synergistic use of Global Navigation Satellite System (GNSS) and Inertial Navigation System (INS) has emerged as the predominant approach to navigate the complex urban environments \cite{godha2007gps}, \cite{li2018high}, \cite{niu2023feature}. Within the domain of GNSS, Real-Time Kinematic (RTK) and Precise Point Positioning (PPP) technologies have been extensively adopted to enhance GNSS/INS integrated solutions. Comparative studies indicate that RTK/INS fusion yields superior accuracy over PPP/INS under identical observational conditions \cite{chen2021ginav}. Nonetheless, the urban environment, with its pervasive obstructions like edifices and arboreal coverage, introduces Non-Line-of-Sight (NLOS) errors to GNSS signals, compromising the accuracy of GNSS/INS integration positioning. Scholars have sought to augment the precision and robustness of GNSS/INS systems in urban canyons by incorporating additional sensory apparatus or by developing techniques to detect and rectify NLOS-induced signal distortions.

Cameras are increasingly utilized in vehicular motion estimation due to their energy efficiency and cost benefits. As an external sensor, cameras can provide rich environmental features for vehicle motion estimation \cite{sun2020plane}, \cite{cheng2019robust}. Consequently, the integration of cameras with GNSS and Micro-Electro-Mechanical System (MEMS)-based Inertial Measurement Units (IMUs) is a common strategy to attain precise localization in complex environments \cite{jiang2021r2}, \cite{shen2024accurate}. Previous research \cite{qin2018vins} introduced Visual-Inertial Navigation Systems (VINS)-monocular model, integrating Visual Inertial Odometry (VIO)-derived relative poses with Global Positioning System (GPS) data within a unified optimization structure. In contrast to VINS-Mono, the work in \cite{liao2021enhancing} combines differential GNSS results with the VIO model, where the VIO model is transformed from the local frame to the global frame, achieving meter-level positioning accuracy in complex urban environments. Advancing from VINS-Mono, the work in \cite{cao2022gvins} introduced the well-known GVINS model, which performs a joint optimization of GNSS pseudorange measurements, visual features, and inertial measurements through factor graph optimization techniques. While methods based on nonlinear optimization have advantages in handling system nonlinearity, multiple iterations of optimization increase computational complexity. Therefore, some researchers have started to focus on Extended Kalman Filter (EKF)-based methodologies. Building on \cite{mourikis2007multi}, the paper \cite{li2019tight} put forth a tightly-coupled Mono/MEMS-IMU/single-frequency GNSS-RTK model employing a Multi-State Constraint Kalman Filter (MSCKF), which attained decimetre-level positioning accuracy in urban environments.

In the above multi-sensor fusion positioning systems, GNSS is the only subsystem that provides absolute position information. Therefore, the quality control of GNSS raw measurements determines the overall performance of the system. This underscores the significance of NLOS signal detection and mitigation, especially in the convoluted terrains of urban environments.

In the detection and mitigation of GNSS NLOS signals, strategies are divided into hardware-centric designs and algorithmic advancements. Compared to expensive hardware improvements such as antenna design in \cite{groves2010novel}, \cite{liu2017compact}, \cite{gupta2016desired}, many researchers have focused on algorithmic improvements. These include empirical weighting models based on elevation angle \cite{won2012weighted}, signal-to-noise ratio (SNR) \cite{groves2013height}, and methods that leverage multi-source information for satellite visibility. Notably, methods augmented by external sources like LiDAR \cite{wen2018exclusion}, \cite{wen2019gnss}, 3-dimensional (3D) maps \cite{wang2013gnss}, \cite{hsu20163d}, and cameras \cite{suzuki2014n}, \cite{wen2019tightly} have refined GNSS NLOS signal detection accuracy. Cameras, especially, present a cost-effective alternative to the high expenses and limited scope of LiDAR, and the necessity for continuously updated 3D map databases. Infrared cameras \cite{meguro2009gps} exhibit varying results for objects at different temperatures, making it easier to distinguish between sky and non-sky areas, which is advantageous for determining the satellite's projection location on the sky-view images. However, compared to regular fish-eye cameras, infrared cameras are more costly. Furthermore, these cameras have not yet seen widespread use in consumer market products like smartphones or vehicle-mounted cameras. Subsequently, many research works began to use sky-pointing fish-eye cameras to capture sky-view images. These images were processed using segmentation algorithms \cite{meguro2009gps}, \cite{cohen2009quantification}, \cite{attia2011counting} to distinguish between sky and non-sky areas. Finally, the satellites received by GNSS receiver were projected onto the sky-view images, facilitating the visualization of GNSS NLOS satellites. As seen from the results in \cite{wen2019tightly} and \cite{wang2023sky}, this approach significantly enhances the performance of SPP/INS positioning in complex urban environments. However, these traditional segmentation algorithms may not adapt well to sky-view images with varying lighting conditions. Furthermore, we have observed that the use of sky-view images for GNSS NLOS detection has not been extended to tightly coupled GNSS/INS/Vision systems. Additionally, there is an absence of comparative performance analysis of sky-view images in different GNSS positioning modes, both domestically and internationally.

We aim to extend the sky-view images aided GNSS NLOS detection and mitigation method (named S-NDM) to the tightly coupled GNSS/INS/Vision system, thereby enhancing vehicle positioning performance in urban canyons. Here we particularly emphasize the progressiveness from \cite{wang2023sky}: (a) different from the previous idea of improving the region growth algorithm, we use the algorithm of neural network to achieve segmentation of sky-view images to adapt to different lighting conditions; (b) the original NLOS signal processing algorithm is only used in tightly coupled SPP/INS framework. In this paper, we extend it to tightly coupled SPP/INS/Vision and RTK/INS/Vision framework. In addition, we evaluate the performance of the algorithm in these two frameworks and verify the practicability of the algorithm. This paper emphasizes the following primary contributions:

1) Adaptive Sky-view Images Segmentation: We introduce an adaptive sky-view images segmentation based on Fully Convolutional Networks (FCN) that can adjust to varying lighting conditions, addressing a key limitation of traditional methods.

2) Integration of Sky-GNSS/INS/Vision: We propose an integrated model that combines GNSS, INS, and Vision. And we extend S-NDM method to this model (named Sky-GVIO), enabling a comprehensive approach to vehicle positioning in challenging urban canyon environments.

3) Performance Evaluation: A comprehensive evaluation of S-NDM's performance is conducted, with a focus on its effectiveness within GNSS pseudorange and carrier phase positioning frameworks, thereby shedding light on its applicability across different GNSS-related integration positioning techniques.

4) Open-Source Sky-view Images Dataset: An open-source repository of sky-view images, including training and testing data, is provided at https://github.com/whuwangjr/sky-view-images , contributing a valuable dataset to the research community and mitigating the lack of available resources in this field.

The reminder of this paper is organized as follows: Section \ref{sec:2} gives an overview of the tightly coupled GNSS/INS/Vision system enhanced by S-NDM. The experimental description and result analysis are introduced in Section \ref{sec:3}. Finally, Section \ref{sec:4} summarizes and concludes the study.

% Following the introduction, the integration method for RTK/MEMS/Vision is described, followed by the introduction of the innovation-based EDM method. Next, the experimental setups and processing strategies for the vehicle-borne experiments are described in detail. Then the experimental results of different weighting strategies of GNSS observations in typical urban situations are investigated, and the performances of the innovation-based EDM and residual-based EDM methods are compared. Finally, the conclusions are provided.

\section{System overview} \label{sec:2}
\begin{figure}[t]
  % \vspace{-1em}
  \centering
  \vspace{-1mm}
  \includegraphics[width=0.48\textwidth]{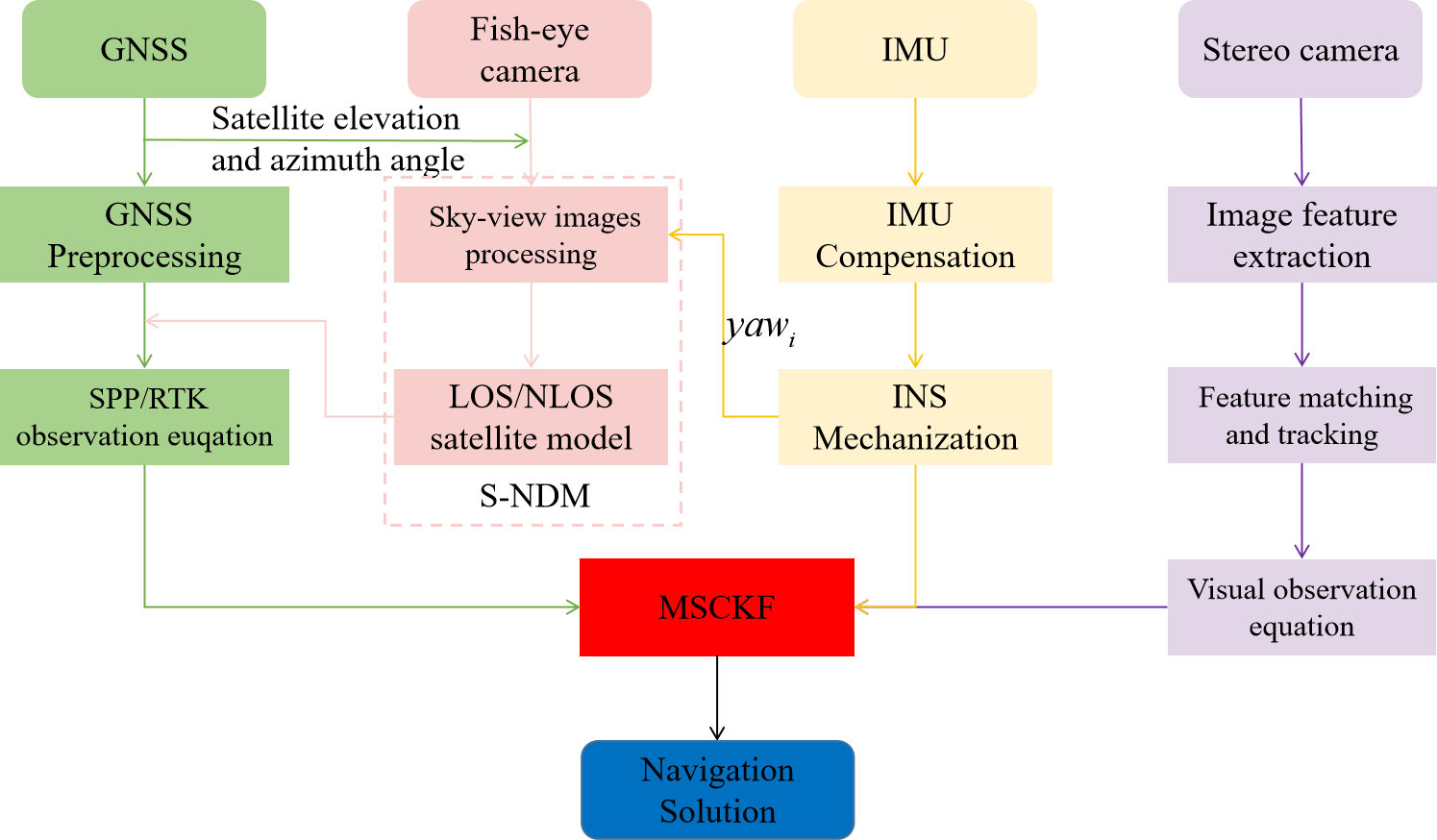}
  \vspace{-3mm}
  \caption{The system structure of the proposed Sky-GVIO.}
  \label{fig:framework}
  \vspace{-4mm}
\end{figure}

The proposed model Sky-GVIO are described in this section, include sky-view images segmentation based on FCN, the tightly coupled GNSS/INS/Vision integration system and S-NDM, as shown Fig. \ref{fig:framework}. The tightly coupled model is a fusion of the observed values. Before the fusion, it is very important to process the GNSS original data. We use S-NDM algorithm to process GNSS NLOS signals. In addition, the INS mechanization is used for state prediction and the system covariance would also be propagated. In the visual part, the feature extraction and tracking will be performed following \cite{vijay2016gray}. Finally, we integrate the observation equations of GNSS, INS and vision into the MSCKF framework to obtain the navigation results. 

\subsection{Sky-view Images Segmentation}
Sky-view images can be significantly affected by factors such as clouds and lighting conditions, making it challenging to achieve high-precision segmentation using traditional methods based on pixel \cite{vijay2016gray}, category \cite{dhanachandra2015image}, region \cite{soltani2020lung}, and so on. It is well-known that FCN represent a mature pixel-level semantic segmentation network \cite{long2015fully}. The FCN network structure primarily consists of two parts: the fully convolutional part and the deconvolution part. The fully convolutional part comprises classical CNN networks, such as VGG and ResNet, which are used for feature extraction. The deconvolution part, on the other hand, upsamples the feature maps to obtain the original-sized semantic segmentation image.

In this paper, the existing ResNet50 \cite{he2016deep} is used for downsampling, which includes 48 convolutional layers. The deconvolution part, on the other hand, upsamples the feature maps to obtain the original-sized semantic segmentation images. In this paper, the upsampling is based on FCN-8s. In the upsampling process, FCN-8s uses transposed convolution to scale the 8x, 16x and 32x feature maps to the original size, and combines these three scaled feature maps by introducing skip connection, so as to ensure the learning of features at different scales.

\begin{figure}[t]
  % \vspace{-1em}
  \centering
  \vspace{-1mm}
  \includegraphics[width=0.48\textwidth]{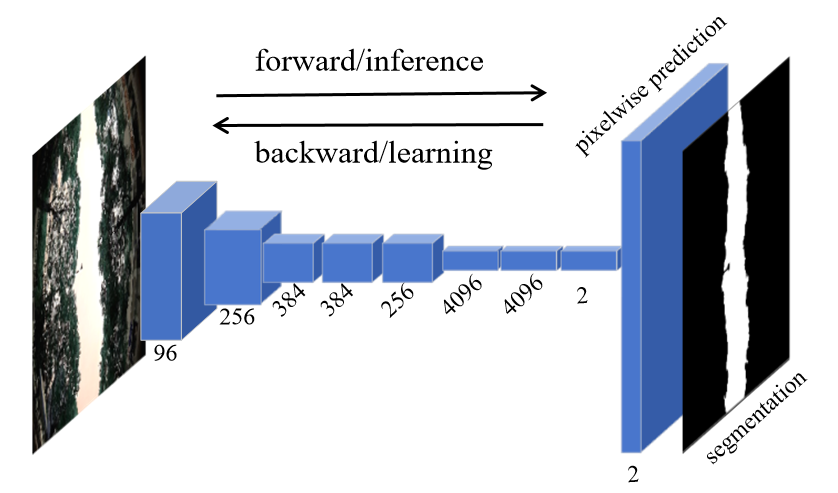}
  \vspace{-3mm}
  \caption{The sky-view images segmentation algorithm based on FCN.}
  \label{fig:fcn}
  \vspace{-4mm}
\end{figure}

The input of FCN can be any size images, the output is the same size as the input, and the number of channels is n (number of target categories) +1 (background). For the sky-view images segmentation task, two types of labels are required (sky region and non-sky region), so the number of channels for sky-view images segmentation algorithm based on FCN is 2. In addition, in this study, we made 440 training datasets by ourselves. As shown in Fig.  \ref{fig:fcn}, we built a sky-view images segmentation model based on FCN.

\subsection{Tightly Coupled GNSS/INS/Vision Integration Model}
GNSS model, INS dynamic model and visual observation model are introduced, respectively. Subsequently, the state model and measurement model of Sky-GVIO integration model are described. Finally, we use the segmentation results to realize the NLOS detection and construct the LOS/NLOS model for NLOS mitigation.

\textbf{1) GNSS Observation Model}: The original pseudorange and carrier phase observation equations in GNSS positioning are expressed as follows:

\begin{equation}
P=\rho+c(t_r-t^s)+I+T+\varepsilon_{p} \label{equ: 1}
\end{equation}
\begin{equation}
L=\rho+c(t_r-t^s)-I+T+\lambda N+\varepsilon_{L}
\end{equation}
where $P$ and $L$ represent the pseudorange and carrier phase, respectively. The angular symbols $s$ and $r$ refer to satellites and receivers, respectively. \(\rho\) denotes the geometric distance between the phase centers of the receiver and satellite antennas. $t_r$ and $t^s$ respectively represent receiver and satellite clock offsets. The speed of light is $c$. $I$ and $T$ refer to the ionospheric and troposphere delay, respectively. $\lambda$ represents the carrier wavelength. $N$ represents carrier phase ambiguity. $\varepsilon_P$ and $\varepsilon_L$ represent pseudorange noise and carrier phase noise, respectively.

For SPP model, equation (\ref{equ: 1}) is sufficient. However, in the case of RTK model, it can be represented as follows:

\begin{small}
\begin{equation}
    \begin{cases} 
    \nabla\Delta P=\nabla\Delta\rho+\nabla\Delta I+\nabla\Delta T+\nabla\Delta\epsilon_P\\ 
    \nabla\Delta L=\nabla\Delta\rho-\nabla\Delta I+\nabla\Delta T+\lambda\nabla\Delta N+\nabla\Delta\epsilon_L \label{equ: 3}
    \end{cases}
\end{equation} 
\end{small}
where $\nabla\Delta$ denotes the double-differenced (DD) operator. The DD operation is used to not only eliminate satellite orbit errors and clock errors but also to mitigate receiver clock errors, tropospheric and ionospheric delays, making it a powerful technique in GNSS positioning.

\textbf{2) INS Dynamic Model}: Considering the noisy measurement of the low-cost IMU, the Coriolis and centrifugal forces due to earth rotation are ignored in the IMU formulation. The inertial measurement can be modeled \cite{xu2023unified} in $b$ (body) frame as follows:

\begin{equation}  
 \tilde{\boldsymbol{a}}_{k} = \boldsymbol{a}_{k} + \boldsymbol{b}_{a_{k}} + \left(\boldsymbol{R}^{n}_{{b}_{k}}\right)^{T}\boldsymbol{g}^{n} + \boldsymbol{n}_{a} \label{equ: 4}
\end{equation}
\begin{equation}  
 \tilde{\boldsymbol{\omega}}_{k} = \boldsymbol{\omega}_{k} + \boldsymbol{b}_{\omega_{k}} + \boldsymbol{n}_{\omega} \label{equ: 5}
\end{equation}
where $\left[\tilde{\boldsymbol{a}}_{k},\tilde{\boldsymbol{\omega}}_{k}\right]$ is the output of the IMU at time $k$ and $\left[\boldsymbol{a}_{k},\boldsymbol{\omega}_{k}\right]$ is the linear acceleration and angular velocity of the IMU sensor. $\boldsymbol{b}_{a_{k}}$ and $\boldsymbol{b}_{\omega_{k}}$ respectively are the biases of the accelerometer and gyroscope at time $k$ . In addition, $\boldsymbol{n}_{a}$ and $\boldsymbol{n}_{\omega}$ are assumed to be zero-mean Gaussian distributed with 
$\boldsymbol{n_a} \sim N\left(0,\Sigma_{n_a}\right)$
, $\boldsymbol{n_{\omega}} \sim N\left(0,\Sigma_{n_{\omega}}\right)$. $\boldsymbol{R}^{n}_{{b}_{k}}$ denotes the rotation matrix from IMU body ($b$)-frame to navigation ($n$)-frame. $\boldsymbol{g}^{n}$ is the gravity in the $n$ frame.

The linearized INS dynamic model \cite{sola2017quaternion} can be expressed as:

\begin{equation}
\begin{cases}
\begin{aligned}
    \delta \dot{\mathbf{p}}^n &= \delta \mathbf{v}^n \\
    \delta \dot{\mathbf{v}}^n &= -\mathbf{R}^n_b\left(\Tilde{\mathbf{a}}-\mathbf{b}_a\right)^{\wedge}
    \delta \bm{\theta}-\mathbf{R}^n_b\delta\mathbf{b}_a-\mathbf{R}_b^n \mathbf{n}_a \\
    \delta \dot{\bm{\theta}} &= -\left(\bm{\Tilde{\omega} - \mathbf{b}}_w\right)^{\wedge}\delta\bm{\theta}-\delta\mathbf{b}_w-\mathbf{n}_w \\
    \delta \dot{\mathbf{b}}_w &= \mathbf{n}_{b_w} \\
    \delta \dot{\mathbf{b}}_a &= \mathbf{n}_{b_a} \\
    \end{aligned} \label{equ: 6}
\end{cases}
\end{equation}
where \(\delta \boldsymbol{\dot{\theta}}\), \(\delta \dot{\mathbf{v}}^n\) and \(\delta \dot{\mathbf{p}}^n\) represent the derivative of attitude, velocity and position errors in \(\mathit{n}\) frame, respectively. The derivatives of \(\delta \dot{\mathbf{b}}_{a}\) and \(\delta \dot{\mathbf{b}}_{\omega}\), denoting the accelerometer and gyroscope biases in \(\mathit{b}\) frame, respectively. In addition, \(\mathbf{R}_{b}^{n}\) represents the rotation matrix from \(\mathit{b}\) frame to \(\mathit{n}\) frame; \(\boldsymbol{\tilde{\omega}}\) and \(\tilde{\boldsymbol{a}}\) represent the outputs of gyroscope and accelerometer, respectively; \(\boldsymbol{b}_{\omega}\) and \(\boldsymbol{b}_{a}\) represent the nominal biases of gyroscope and accelerometer, respectively; \(\delta\boldsymbol{\theta}\) and \(\delta\mathbf{v}^n\) represent the errors of attitude and velocity in \(\mathit{n}\) frame, respectively; \(\delta \boldsymbol{b}_{\omega}\) and \(\delta \boldsymbol{b}_{a}\) represent the errors of gyroscope bias and accelerometer bias, respectively; \(\boldsymbol{n}_{\omega}\) and \(\boldsymbol{n}_{a}\) represent the noises of angular rate and acceleration, respectively; \(\mathbf{n}_{b_{a}}\) and \(\mathbf{n}_{b_{\omega}}\) represent the noises of gyroscope bias and accelerometer bias, respectively. The symbol \(\left(\cdot\right)^{\wedge}\) is the cross-product.

Therefore, the error state vector of INS can be expressed as:

\begin{equation}
\delta\boldsymbol{x}_{ins}=\left[\delta\mathbf{p}^{n}\quad\delta\mathbf{v}^{n}\quad\delta\boldsymbol{\theta}\quad\delta\mathbf{b}_{a}\quad\delta\mathbf{b}_{\omega}\right]^{T} \label{equ: 7}
\end{equation}

\textbf{3) Visual Measurement Model}: The core idea of the well-known MSCKF is to establish geometric constraints between multi-camera states by utilizing the same visual feature points observed by multi-cameras. Following this concept, we establish a visual model. For a visual feature point $f^j$ observed by a stereo camera at time $i$, its visual observation model \cite{xu2023unified} on the normalized projection planes of the left and right cameras can be represented as follows:
\begin{equation}
\mathit{z}_{cam,i}^{j}=\left[\begin{matrix}\mathit{u}_{c_{0,i}}^{j}\\\mathit{v}_{c_{0,i}}^{j}\\\mathit{u}_{c_{1,i}}^{j}\\\mathit{v}_{c_{1,i}}^{j}\end{matrix}\right]=\left[\begin{matrix}\frac{1}{\mathit{Z}_{c_{0,i}}^{j}}\mathbf{I}_{2\times2}&\mathbf{0}_{2\times2}\\\mathbf{0}_{2\times2}&\frac{1}{\mathit{Z}_{c_{1,i}}^{j}}\mathbf{I}_{2\times2}\end{matrix}\right]\left[\begin{matrix}\mathit{X}_{c_{0,i}}^{j}\\\mathit{Y}_{c_{0,i}}^{j}\\\mathit{X}_{c_{1,i}}^{j}\\\mathit{Y}_{c_{1,i}}^{j}\end{matrix}\right]+\mathit{\epsilon}_{cam,i}^{j}
\end{equation}
where the subscripts 0 and 1 represent the left and right cameras, respectively. \(\left(\mathit{u}_{c_{0,j}}^{j},\mathit{v}_{c_{0,j}}^{j}\right)^{T}\) and \(\left(\mathit{u}_{c_{1,j}}^{j},\mathit{v}_{c_{1,j}}^{j}\right)^{T}\) are the pixel coordinates of the same feature point on the normalized plane for left camera and right camera, respectively. \(\mathit{\epsilon}_{cam,i}^{j}\) is visual measurement noise. \(\left(\mathit{X}_{c_{0,j}}^{j},\mathit{Y}_{c_{0,j}}^{j},\mathit{Z}_{c_{0,j}}^{j}\right)^{T}\) and \(\left(\mathit{X}_{c_{1,j}}^{j},\mathit{Y}_{c_{1,j}}^{j},\mathit{Z}_{c_{1,j}}^{j}\right)^{T}\) represents the position of the same feature point for left camera and right camera in \(\mathit{c}\) frame, which can be expressed as:

\begin{equation}
\begin{bmatrix}\mathit{X}_{c_{0,i}}^{j}\\\mathit{Y}_{c_{0,i}}^{j}\\\mathit{Z}_{c_{0,i}}^{j}\end{bmatrix}=\left(\mathbf{R}_{c_{0,i}}^{n}\right)^{T}\left(\boldsymbol{p}_{j}^{n}-\boldsymbol{p}_{c_{0,i}}^{n}\right)
\end{equation}

\begin{equation}
\begin{bmatrix}\mathit{X}_{c_{1,i}}^{j}\\\mathit{Y}_{c_{1,i}}^{j}\\\mathit{Z}_{c_{1,i}}^{j}\end{bmatrix}=\left(\mathbf{R}_{c_{0,i}}^{c_{1,i}}\right)^{T}\left(\boldsymbol{p}_{j}^{c_{0,i}}-\boldsymbol{p}_{c_{1,i}}^{c_{0,i}}\right)
\end{equation}
where \(\mathbf{R}_{c_{0,i}}^{c_{1,i}}\) and \(\boldsymbol{p}_{c_{0,i}}^{n}\) are the rotation matrix and position of the left camera at time \(\mathit{i}\) in \(\mathit{n}\) frame, respectively. \(\mathbf{R}_{c_{0,i}}^{c_{1,i}}\) is the rotation matrix from left camera to right camera at time \(\mathit{i}\), \(\boldsymbol{p}_{c_{1,i}}^{c_{0,i}}\) is the translation matrix from left camera to right camera, which can be accurately corrected in advance \cite{furgale2013unified}. \(\boldsymbol{p}_{j}^{n}\) and \(\boldsymbol{p}_{j}^{c_{0,i}}\) respectively are the positions of the same visual feature point in \(\mathit{n}\) frame and left \(\mathit{c}\) frame.

We adopted the method proposed by \cite{sun2018robust} to construct the visual reprojection error between relative camera poses, and the visual state vector was described as:

\begin{equation}
\delta\boldsymbol{x}_{cam} = \left[\delta\boldsymbol{\theta}^n_{c_1}\quad\delta\boldsymbol{p}^n_{c_1}\quad\delta\boldsymbol{\theta}^n_{c_2}\quad\delta\boldsymbol{p}^n_{c_2}\quad...\quad\delta\boldsymbol{\theta}^n_{c_s}\quad\delta\boldsymbol{p}^n_{c_s}\right]^{T} \label{equ: 11}
\end{equation}
where \(\delta\boldsymbol{\theta}^n_{c_i}\) and \(\delta\boldsymbol{p}^n_{c_i}\) are attitude errors and position errors at time \(\mathit{i}\). The subscript \(\mathit{s}\) represents the total number of camera poses in the sliding window. The measurement equation of visual reprojection error is expressed as follows:

\begin{equation}
\delta \boldsymbol{z}_{cam} = \boldsymbol{\tilde{z}}_{cam} - \boldsymbol{\hat{z}}_{cam} = \boldsymbol{H}_{cam}\delta\boldsymbol{x}_{cam} + \boldsymbol{V}_{cam} \label{equ: 12}
\end{equation}
\(\boldsymbol{\tilde{z}}_{cam}\) and \(\boldsymbol{\hat{z}}_{cam}\) represent visual observations and visual reprojection observations, respectively; \(\boldsymbol{H}_{cam}\) represents the Jacobi matrix of stereo camera positioning model.

\textbf{4) State and Measurement model of the Tightly Coupled GNSS/INS/Vision}: This paper employs MSCKF for the tightly coupled GNSS/INS/Vision integration. Based on the above introductions of different sensor models, the complete state model for the tightly coupled GNSS/INS/Vision integration is as follows:

\begin{equation}
\delta\boldsymbol{x}=\left[\delta\boldsymbol{x}_{ins}\quad\delta\boldsymbol{x}_{GNSS}\quad\delta\boldsymbol{x}_{cam}\right]^{T}
\end{equation}

For both SPP and RTK positioning modes, this paper has constructed state models separately:

\begin{equation}
    \delta x_{GNSS,SPP}=[\delta t_r]^T \label{equ: 14}
\end{equation}
\begin{equation}
    \delta x_{GNSS,RTK}=[\delta\nabla\Delta N]^T \label{equ: 15}
\end{equation}
where $\delta\nabla\Delta N$  represents the DD carrier phase ambiguity.In addition, the error state model of INS and vision have already been provided in equation (\ref{equ: 7}) and (\ref{equ: 11}).

The state prediction model for the tightly coupled GNSS/INS/Vision integration is as follows:

\begin{equation}
\begin{small}
\begin{bmatrix}\delta\boldsymbol{\dot{x}}_{ins}\\\delta\boldsymbol{\dot{x}}_{GNSS}\\\delta\boldsymbol{\dot{x}}_{cam}\end{bmatrix}=\begin{bmatrix}\boldsymbol{F}_{ins}&\mathbf{0}&\mathbf{0}\\\mathbf{0}&\mathbf{0}&\mathbf{0}\\\mathbf{0}&\mathbf{0}&\mathbf{0}\end{bmatrix}\begin{bmatrix}\delta\boldsymbol{x}_{ins}\\\delta\boldsymbol{x}_{GNSS}\\\delta\boldsymbol{x}_{cam}\end{bmatrix}+\begin{bmatrix}\boldsymbol{n}_{ins}\\\boldsymbol{n}_{GNSS}\\\mathbf{0}\end{bmatrix}
\end{small}
\end{equation}
where \(\boldsymbol{F}_{ins}\) is the system matrices of INS state which could be directly from equation (\ref{equ: 6}). \(\boldsymbol{n}_{ins}\) and \(\boldsymbol{n}_{GNSS}\) are the process noises of INS and map, respectively. In addition, the camera poses in the sliding window are considered constant, so its process noise is \(\mathbf{0}\). Based on equation (\ref{equ: 6}), the special form of  can be written as:

\begin{equation}
\boldsymbol{F}_{ins}=\begin{bmatrix}\mathbf{0}&{\mathbf{I}}&\mathbf{0}&\mathbf{0}&\mathbf{0}\\\mathbf{0}&\mathbf{0}&-\mathbf{R}^n_b\left(\Tilde{\mathbf{a}}-\mathbf{b}_a\right)^{\wedge}&-\mathbf{R}^n_b&\mathbf{0}\\\mathbf{0}&\mathbf{0}&-\left(\bm{\Tilde{\omega} - \mathbf{b}}_w\right)^{\wedge}&\mathbf{0}&-\mathbf{I}\\\mathbf{0}&\mathbf{0}&\mathbf{0}&\mathbf{0}&\mathbf{0}\\\mathbf{0}&\mathbf{0}&\mathbf{0}&\mathbf{0}&\mathbf{0}\end{bmatrix} \label{equ: 17}
\end{equation}
where $\mathbf{I}$ is the identity matrix.

To deal with discrete time measurement from the INS, the 4th order Runge-kutta [7] numerical integration of equation (\ref{equ: 17}) to propagate the estimated state variables. It is worth noting, only the IMU state variables are propagated, the visual and GNSS state variables are only copied. Meanwhile, we also need to propagate the covariance of the state:

\begin{equation}
\begin{aligned}
\boldsymbol{P}_{k,k-1} &= \boldsymbol{\Phi}_{k,k-1} \boldsymbol{P}_{k-1} \boldsymbol{\Phi}_{k,k-1}^T + \boldsymbol{Q}_{k-1}\\
\boldsymbol{\Phi}_{k,k-1} &= \boldsymbol{\Phi}(t_{k-1}, t_k) = \exp(\int_{t_k}^{t_{k-1}} \boldsymbol{F}(\tau)\mathrm{d}\tau)
\end{aligned}
\end{equation}
where $\boldsymbol{\Phi}_{k,k-1}$ represents the discrete state transition matrix, $\boldsymbol{F}(\tau)$ is the continuous time state transition matrix at time $\tau$ ($\tau \in \left(t_k, t_{k+1}\right)$) and $\boldsymbol{Q}_{k-1}$ is the discrete time noise covariance. $\boldsymbol{P}_{k-1}$ represents the error state covariance matrix before augmentation. $\boldsymbol{P}_{k,k-1}$ represents the one-step prediction error covariance matrix from time $t_{k-1}$ to time $t_k$.

It is worth noting that every time a new image is recorded, the state and covariance matrix will be augmented with a copy of the current camera pose estimate. The initial value of camera pose is derived from the INS mechanization and the covariance matrix $\boldsymbol{P}_k$ after augmented can be expressed as:

\begin{equation}
\boldsymbol{P}_k=\left[\begin{matrix} \mathbf{I}_{15+y+6m} \\ \mathbf{J}\end{matrix}\right] \boldsymbol{P}_{k-1} \left[\begin{matrix} \mathbf{I}_{15+y+6m} \\ \mathbf{J}\end{matrix}\right]^T
\end{equation}

where $y$ and $m$ represent the number of variables related to GNSS and vision at a certain moment. And when the GNSS is recorded, we only need to remove and add state variables and corresponding covariance. $\mathbf{J}$ is the Jacobi matrix, which has the following form:

\begin{equation}
\mathbf{J}=\left[\begin{matrix}
(\mathbf{R}_c^b)^T&\mathbf{0}&\mathbf{0}&\mathbf{0}&\mathbf{0}&\mathbf{0}_{3\times(y+6m)}\\-\mathbf{R}_b^c(\boldsymbol{p}_c^b)^{\wedge}&\mathbf{0}&\mathbf{I}&\mathbf{0}&\mathbf{0}&\mathbf{0}_{3\times(y+6m)}    
\end{matrix}\right]
\end{equation}
where $\mathbf{R}_c^b$ and $\boldsymbol{p}_c^b$ are the rotation matrix and translation matrix between camera and IMU, which are calibrated offline [36].

Based on the previous equations, the measurement equation for the tightly coupled SPP/INS/Vision integration are formulated as follows:

\begin{equation}
    \begin{aligned}
\left[\begin{matrix}\delta \boldsymbol{P}_{SPP} \\ \delta \boldsymbol{z}_{cam}\end{matrix}\right]&=\left[\begin{matrix}{\bf H}_{P,SPP} \\ {\bf H}_{cam}\end{matrix}\right]\left[\begin{matrix}\delta \boldsymbol{x}_{ins} \\ \delta \boldsymbol{x}_{GNSS,SPP} \\ \delta \boldsymbol{x}_{cam}\end{matrix}\right]+\left[\begin{matrix}\boldsymbol{\varepsilon}_{P,SPP} \\ \boldsymbol{\varepsilon}_{cam}\end{matrix}\right]\\
\left[\begin{matrix}\delta \boldsymbol{P}_{SPP} \\ \delta \boldsymbol{z}_{cam}\end{matrix}\right]&=\left[\begin{matrix}\boldsymbol{P}-\boldsymbol{\hat{P}}_{ins} \\ \boldsymbol{z}_{cam}-\hat{\boldsymbol{z}}_{cam}\end{matrix}\right] \label{equ: 21}
    \end{aligned}
\end{equation}
where $\delta \boldsymbol{P}_{SPP}$ is error of pseudorange observation in SPP and $\delta \boldsymbol{z}_{cam}$ is error of visual observation in equation (\ref{equ: 12}). ${\bf H}_{P,SPP}$ is the Jacobi matrix of pseudorange error and ${\bf H}_{cam}$ is the Jacobi matrix of the involved camera states in equation (\ref{equ: 12}). Then $\delta \boldsymbol{x}_{ins}$, $\delta \boldsymbol{x}_{GNSS,SPP}$ and $\delta \boldsymbol{x}_{cam}$ are the error state vectors of INS, SPP and Vision which can be found in equation (\ref{equ: 7}), (\ref{equ: 14}) and (\ref{equ: 11}), respectively. In the same way, $\boldsymbol{\varepsilon}_{P,SPP}$ and $\boldsymbol{\varepsilon}_{cam}$ denote pseudorange observation error noise in SPP and visual observation error noise, respectively. In addition, $P$ and $\boldsymbol{\hat{P}}_{ins}$ are the actual measured pseudorange in equation (\ref{equ: 1}) and the pseudorange predicted by INS mechanization, respectively. $\boldsymbol{z}_{cam}$ and $\hat{\boldsymbol{z}}_{cam}$ respectively represent the observed and reprojected visual measurements in equation (\ref{equ: 12}).

The measurement equation for the tightly coupled RTK/INS/Vision integration are formulated as follows:

\begin{equation}
\begin{small}
\begin{aligned}
\left[\begin{matrix}\delta \boldsymbol{P}_{RTK} \\ \delta \boldsymbol{L}_{RTK}\\\delta \boldsymbol{z}_{cam}\end{matrix}\right]&=\left[\begin{matrix}{\bf H}_{P,RTK} \\ {\bf H}_{L,RTK}\\{\bf H}_{cam}\end{matrix}\right]\left[\begin{matrix}\delta \boldsymbol{x}_{ins} \\ \delta \boldsymbol{x}_{GNSS,RTK} \\ \delta \boldsymbol{x}_{cam}\end{matrix}\right]+\left[\begin{matrix}\boldsymbol{\varepsilon}_{\nabla\Delta P,RTK} \\ \boldsymbol{\varepsilon}_{\nabla\Delta L,RTK}\\\boldsymbol{\varepsilon}_{cam}\end{matrix}\right]\\
\left[\begin{matrix}\delta \boldsymbol{P}_{RTK} \\ \delta \boldsymbol{L}_{RTK}\\ \delta \boldsymbol{z}_{cam}\end{matrix}\right]&=\left[\begin{matrix}\boldsymbol{\nabla\Delta P}-\boldsymbol{\nabla\Delta\hat{P}}_{ins} \\ \boldsymbol{\nabla\Delta L}-\boldsymbol{\nabla\Delta\hat{L}}_{ins}\\\boldsymbol{z}_{cam}-\hat{\boldsymbol{z}}_{cam}\end{matrix}\right] \label{equ: 22}
\end{aligned}
\end{small}
\end{equation}
where $\delta \boldsymbol{P}_{RTK}$ and $\delta \boldsymbol{L}_{RTK}$ represent the observation errors of DD pseudorange and DD carrier phase in RTK, respectively. $\boldsymbol{\nabla\Delta P}$ and $\boldsymbol{\nabla\Delta L}$ can be found in equation (\ref{equ: 3}). In addition, $\boldsymbol{\nabla\Delta\hat{P}}_{ins}$ and $\boldsymbol{\nabla\Delta\hat{L}}_{ins}$ are DD pseudorange predicted and DD carrier phase predicted by INS mechanization, respectively. Then ${\bf H}_{P,RTK}$ and ${\bf H}_{L,RTK}$ are the Jacobi matrices of DD pseudorange error and DD carrier phase error. $\delta \boldsymbol{x}_{GNSS,RTK}$ is the error state vector of RTK which can be found in equation (\ref{equ: 15}). $\boldsymbol{\varepsilon}_{\nabla\Delta P,RTK}$ and $\boldsymbol{\varepsilon}_{\nabla\Delta L,RTK}$ denote DD pseudorange observation error noise and DD carrier phase observation error in RTK. 

\begin{equation}
\begin{small}
\begin{aligned}
\begin{cases}
R_k^{P,LOS}=f\times(10^{\frac{SNR-S_1}{a}}(( \frac{A}{\frac{S_0-S_1}{10} } -1)\frac{SNR-S_1}{S_0-S_1}+1))\times\sigma_p^2\\
R_k^{L,LOS}=f\times(10^{\frac{SNR-S_1}{a}}(( \frac{A}{\frac{S_0-S_1}{10} } -1)\frac{SNR-S_1}{S_0-S_1}+1))\times\sigma_L^2 \label{equ: 23}
\end{cases} 
\end{aligned}
\end{small}
\end{equation}

\begin{equation}
\begin{aligned} 
\begin{cases}
R_k^{P,NLOS}&=K \times R_k^{P,LOS}\\
R_k^{L,NLOS}&=K \times R_k^{L,LOS}  \label{equ: 24}
\end{cases}
\end{aligned}  
\end{equation}
In equation (\ref{equ: 23}), $f={1}/{\sin^2(ele)}$, $ele$ and $SNR$ refer to elevation angles and SNR of satellites, respectively. The work \cite{herrera2016gogps} gives $s_1=50$, $A=30$, $s_0=10$ and $a=20$, these parameters are empirical values. $\sigma_P$ and $\sigma_L$ represent the standard deviation of pseudorange and carrier phase respectively, which are 0.3 m and 0.03 m given in this paper. The $K$ is the scale factor, which is 10 in this paper. If the satellite's projection is located in the sky semantic region, then Stochastic model of satellite observations will be modeled by using equation (\ref{equ: 24}), and if not, it will be modeled by using equation (\ref{equ: 23}). 

\subsection{The Sky-view Images aided GNSS NLOS Detection and Mitigation Method (S-NDM)}
\begin{figure}[t]
  % \vspace{-1em}
  \centering
  \vspace{-1mm}
  \includegraphics[width=0.48\textwidth]{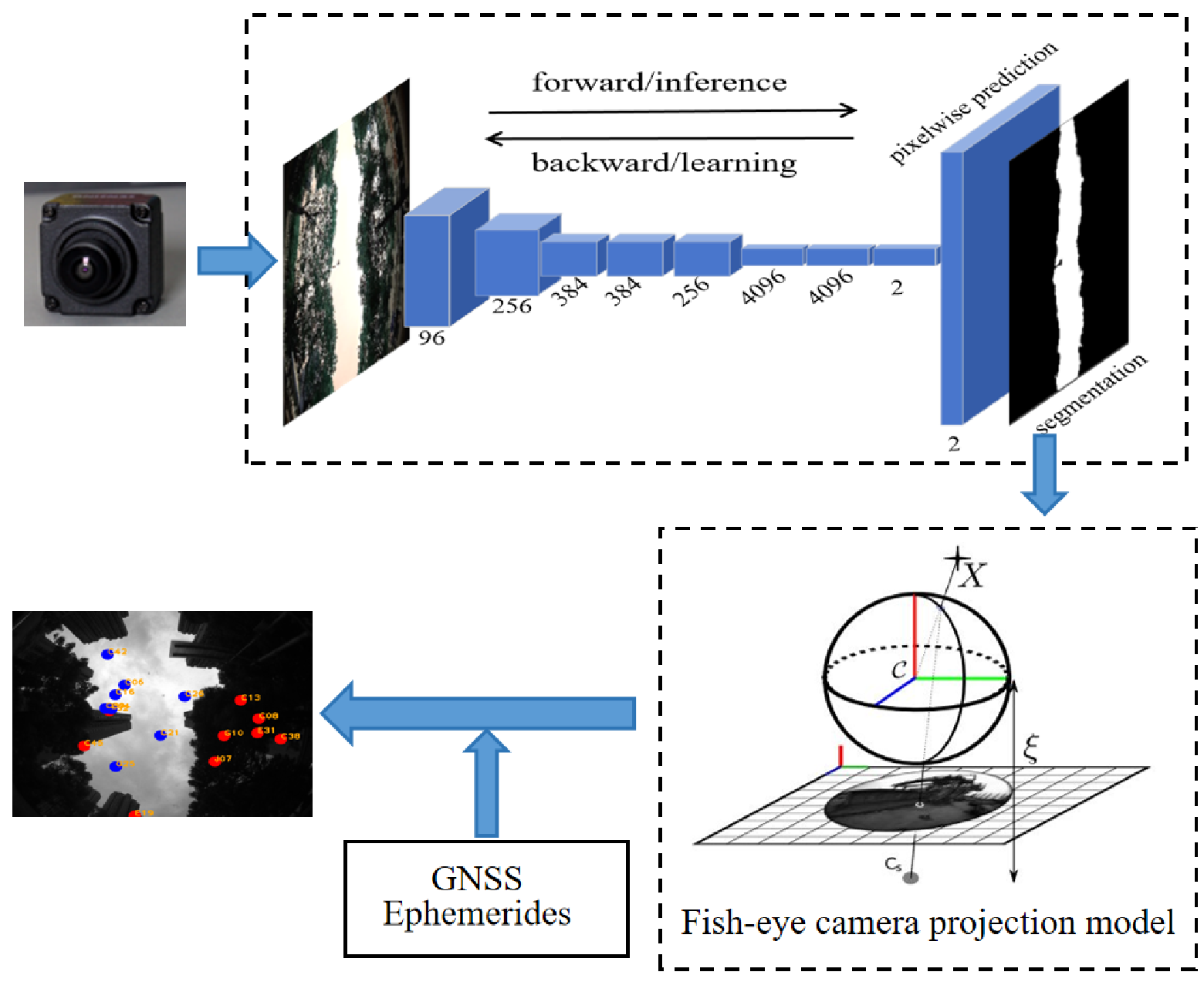}
  \vspace{-3mm}
  \caption{The overall flow of S-NDM algorithm.}
  \label{fig:s-ndm}
  \vspace{-4mm}
\end{figure}

For accurate modeling of the GNSS noise covariance in the tightly coupled GNSS/INS/Vision integration, it is essential to differentiate between LOS and NLOS satellites. Therefore, we obtain sky-mask after performing semantic segmentation of the sky-view images using FCN, and subsequently, based on the projection model mentioned in \cite{wang2023sky} and satellite information included elevation and azimuth angles provided by satellite ephemeris, we ultimately identify LOS and NLOS conditions around the GNSS receiver. Fig. \ref{fig:s-ndm} shows the overall flow of S-NDM algorithm. In this process, if the satellite's projection is located in the sky region, then the satellite will be classified as an LOS satellite (represented by a blue dot), and if not, it will be classified as an NLOS satellite (represented by a red dot). Through this satellite visualization strategy, we can obtain the judgment conditions of equation (\ref{equ: 23}) and (\ref{equ: 24}). Different from judging conditions by experience threshold in \cite{herrera2016gogps}, satellite visualization strategy is more reliable, which is also the difference between LOS/NLOS signal modeling in this paper and traditional methods.

\begin{table*}[t!]
	\captionsetup {font={small,stretch=1.5}}
	\caption{Technical specifications of the IMU sensors.} 
	\label{tab: 1}
	\begin{center}
	\setlength\tabcolsep{15pt}
	\setlength{\belowcaptionskip}{1pt}
    \renewcommand\arraystretch{1.0}
	\small
 	\vspace{-2.0em}
	\resizebox{\linewidth}{!}{
		\begin{tabular}{ccccccc}
			\toprule
			  \multirow{2}{*}{IMU Equipment} & \multirow{2}{*}{Grade} & \multirow{2}{*}{Sample rates (Hz)} & Angular & Velocity & Acc & Gyro\\
             &  &  & $\left(^\circ/\sqrt{h}\right)$ & $\left(m/s/\sqrt{h}\right)$ & $\left(mGal\right)$ & $\left(^\circ/h\right)$\\
			\midrule
	ADIS-16470  & MEMS & 100 & 0.34 & 0.18 & 1300 & 8 \\
        SPAN-ISA-100C  & Tactical & 200 & 0.005 & 0.018 & 100 & 0.05  \\
             \specialrule{0.08em}{1pt}{1pt}
		\end{tabular}
		\vspace{-2em}
		}
	\end{center}
\end{table*}

\section{Experiments} \label{sec:3}
This section delineates the experimental methodology undertaken to assess the effectiveness of the sky-view image-assisted GNSS NLOS detection across different positioning models. The study categorizes the models into SPP-related and RTK-related tightly coupled models for comparative analysis. To evaluate positioning performance, we calculated the root mean square error (RMSE) in the three directions of the East (E), North (N) and Up (U).
\subsection{Experiment Description}
\begin{figure}[t]
  % \vspace{-1em}
  \centering
  \vspace{-1mm}
  \includegraphics[width=0.48\textwidth]{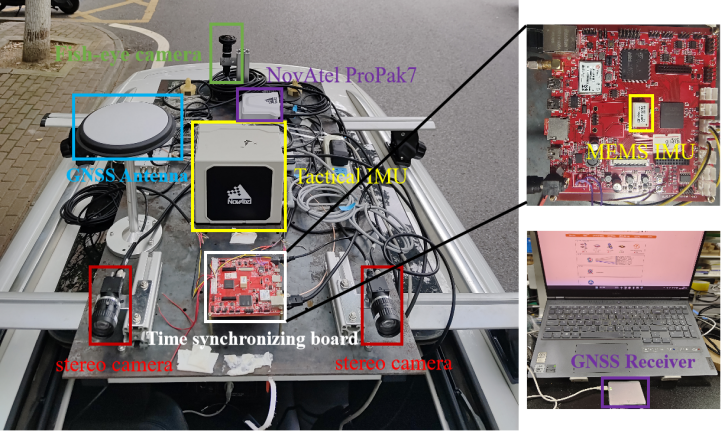}
  \vspace{-3mm}
  \caption{Illustration of experimental hardware platform.}
  \label{fig:hardware}
  \vspace{-4mm}
\end{figure}

As is shown in Fig. \ref{fig:hardware}, The data acquisition platform consists of a GNSS receiver (Septentrio mosaic-X5 mini), GNSS antenna (NovAtel GNSS-850), and two forward-looking cameras (FLIR BFS-U3-31S4C-C), an sky-pointing fish-eye camera (FE185C057HA-1), a tactical grade IMU (NovAtel SPAN-ISA-100C), a MEMS-IMU (ADIS-16470), and a time synchronization board. The time synchronization board unifies the time of all sensors to GPS time through pulse per second (PPS) generated by the GNSS receiver. The sampling rates of GNSS, MEMS-IMU, forward-looking cameras and fish-eye camera are 1 Hz, 100 Hz, 10 Hz, and 1 Hz, respectively. In addition, the NovAtel SPAN-ISA-100C interacts with NovAtel's ProPak7 receiver via a highly reliable IMU interface. The tightly coupled multi-GNSS post-processing kinematic (PPK)/INS bidirectional smoothing position results can be obtained through commercial IE 8.9 software and used as a reference truth value. Table \ref{tab: 1} lists the specific parameters of the two IMUs. For software, we run the Linux system in the environment with Intel Core i7-9750H@ 2.6GHz, 32GB memory. In addition, we used a 3060Ti GPU (Graphics Processing Unit) for acceleration. Meanwhile, we used the opencv3.4.9 \cite{bradski2000opencv} to process the images in the tightly coupled system.

\begin{figure}[t]
  % \vspace{-1em}
  \centering
  \vspace{-1mm}
  \includegraphics[width=0.48\textwidth]{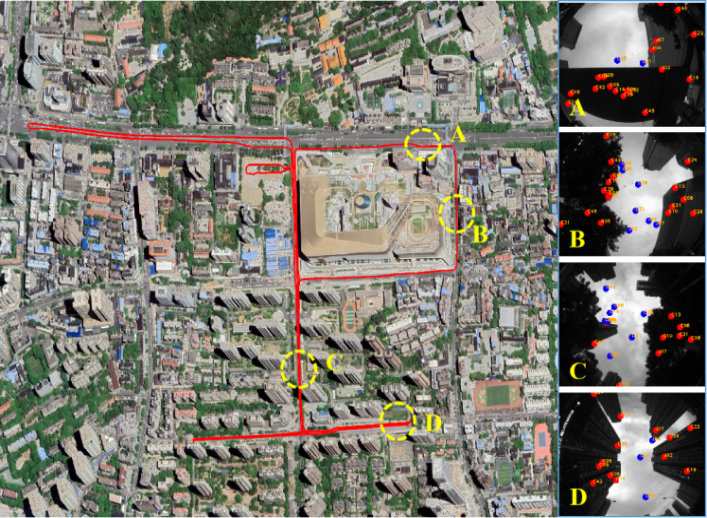}
  \vspace{-3mm}
  \caption{The experimental route and scene in the urban canyon. (A, B, C and D on the right correspond to the sky-view images of the four scenes in the trajectory, respectively. In the sky-view images on the right, the red dots represent the NLOS satellite, the blue dots represent the LOS satellite.)}
  \label{fig:route}
  \vspace{-4mm}
\end{figure}

\begin{figure}[t]
  % \vspace{-1em}
  \centering
  \vspace{-1mm}
  \includegraphics[width=0.48\textwidth]{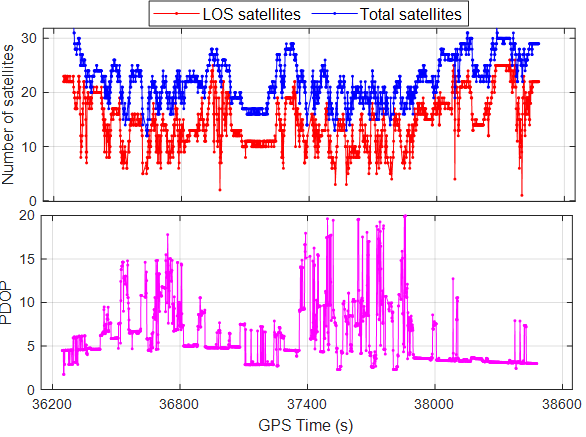}
  \vspace{-3mm}
  \caption{Number of LOS and total satellites (top) and PDOP value of all satellites (bottom).}
  \label{fig:pdop}
  \vspace{-4mm}
\end{figure}

We collected vehicular data in a typical urban canyon area in Wuhan On September 3, 2023. The experimental trajectory and surrounding landscape, featuring high-rise structures, dense foliage, and overpasses, are illustrated in Fig. \ref{fig:route}. The GNSS elevation angles and the position dilution of precision (PDOP) values for this route are shown in Fig. \ref{fig:pdop}. Combined with the LOS/NLOS satellite conditions presented in Fig. \ref{fig:route}, it is conceivable that our testing environment is plagued by severe GNSS NLOS, multipath, and cycle slip issues. These complications not only impair GNSS-based positioning accuracy but also challenge the reliability of GNSS/INS/Vision integration model that depends on GNSS for absolute position information. The choice of such a challenging environment underscores the purpose of the experiment, which is to validate the efficacy and dependability of S-NDM algorithm and Sky-GVIO model introduced in this study.

\begin{figure*}[hptb]
	    \centering
     \vspace{-1mm}
	    \includegraphics[width=2\columnwidth]{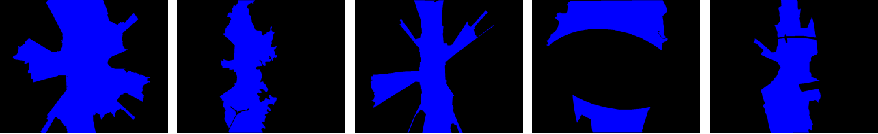}
      \vspace{-3mm}
	    \caption{The examples of sky-view images dataset by annotated semantically.}	\label{fig:annotated}
     \vspace{-4mm}
\end{figure*}

In addition, we briefly introduce our sky-view images dataset, including the training dataset and the testing dataset. The training dataset contains 440 images and the testing dataset contains 2000 images. These data were collected in typical urban canyons in two different areas of Wuhan, which contain trees, tall buildings, and light poles. Therefore, this dataset is very suitable for sky-view images segmentation experiments. As shown in Fig. \ref{fig:annotated}, it is the sky-view images that we have labeled semantically, with blue representing the sky area and black representing the non-sky area.

\subsection{The Results of Sky-view Images Segmentation and GNSS NLOS Detection}
\begin{figure*}[hptb]
	    \centering
     \vspace{-1mm}
	    \includegraphics[width=2\columnwidth]{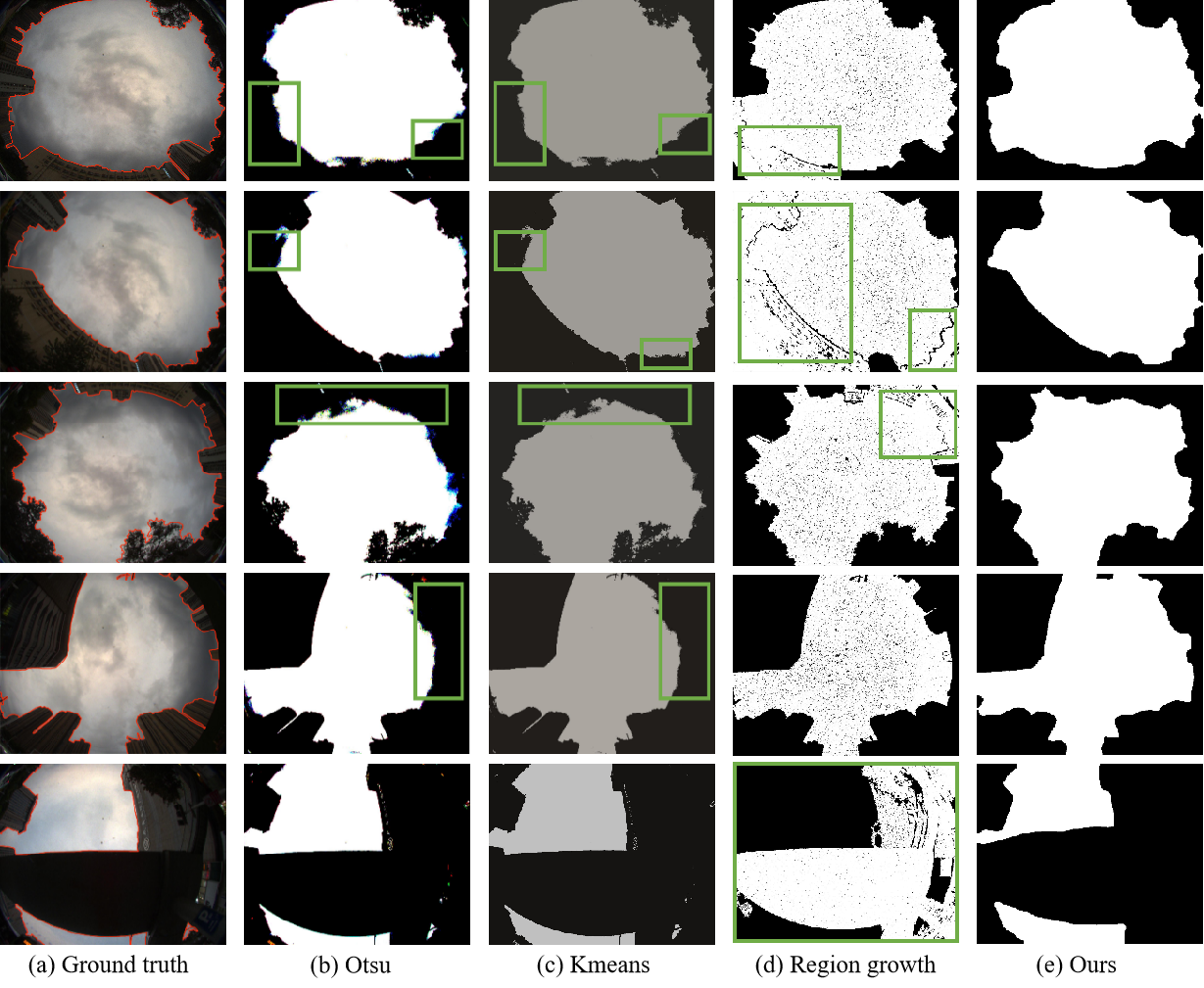}
      \vspace{-3mm}
	    \caption{Experimental results of sky-view images segmentation using different methods. From left to right: Ground truth with the outline of sky-region (red line), the segmentation results by Otsu, Kmeans, Region growth and our proposed method, respectively. Regions that are incorrectly segmented  are highlighted via green boxes.}	
     \label{fig:four methods}
     \vspace{-4mm}
\end{figure*}

The segmentation of sky-view images in urban canyons is challenging due to dynamic environmental factors such as cloud cover and varying light conditions, which can degrade the accuracy of traditional image segmentation techniques. The Result of poor image segmentation accuracy will lead to errors in NLOS detection. Fig. \ref{fig:four methods} presents a comparison of the results of sky-view images segmentation between traditional segmentation algorithms and the method proposed in this paper.

We compare our method based FCN with representative methods on image segmentation, including Otsu, Kmeans and Region growth. For fair comparison with the other competitors, all tests were performed on our collected dataset. As can be seen in Fig. \ref{fig:four methods}, cloud and light cause obvious errors in Fig. \ref{fig:four methods}(b), Fig. \ref{fig:four methods}(c) and Fig. \ref{fig:four methods}(d), especially in areas close to buildings. The performances of Otsu and Kmeans are relatively similar, but Region Growth demonstrates a higher incidence of misclassification. Furthermore, in images featuring elevated bridges, an erroneous selection of seed points leads to the misidentification of sky regions as non-sky regions, which can be critically detrimental in NLOS identification. In contrast, our proposed FCN-based approach attains a high-precision segmentation outcome. This is because FCN captures global context information for the input image by using convolution and pooling layers. This allows the network to better understand the relationships between different objects and the overall structure of the image, which helps in more accurate segmentation.

\begin{figure}[t]
  % \vspace{-1em}
  \centering
  \vspace{-1mm}
  \includegraphics[width=0.48\textwidth]{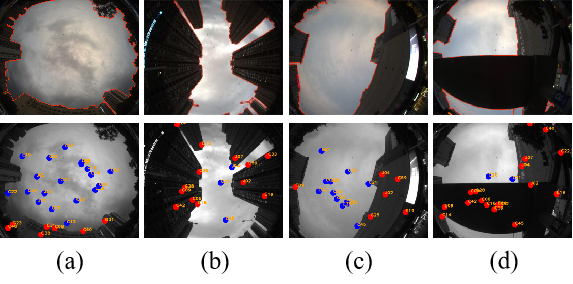}
  \vspace{-3mm}
  \caption{The detection visibility of LOS/NLOS satellites. (The red dots represent the NLOS satellite, the blue dots represent the LOS satellite.)}
  \label{fig:visibility}
  \vspace{-4mm}
\end{figure}

FCN-derived segmentation results are utilized for NLOS detection, with the visibility of satellites illustrated in Fig. \ref{fig:visibility}. In the Fig. \ref{fig:visibility}, we can observe that there are no identification errors in the visualization results of LOS and NLOS satellites, underscoring the reliability of our S-NDM algorithm.
In the case of mild urban canyon environments (as illustrated in Fig. \ref{fig:visibility}(a) and Fig. \ref{fig:visibility}(c)), LOS satellites dominate. However, in the case of deep urban canyons (as shown in Fig. \ref{fig:visibility}(b)) and environments with elevated bridges (as in Fig. \ref{fig:visibility}(d)), fewer than four LOS satellites are detectable, and the satellite configurations are suboptimal, highlighting the complexity and challenges of our experimental testing environments.

\begin{table}[t!]
	\captionsetup {font={small,stretch=5.5}}
	\caption{Performance comparison of sky-view images using different methods} 
	\label{tab: 2}
	\begin{center}
	\setlength\tabcolsep{10pt}
	\setlength{\belowcaptionskip}{1pt}
    \renewcommand\arraystretch{1.0}
	\small
 	\vspace{-2.0em}
	\resizebox{\linewidth}{!}{
		\begin{tabular}{c|cccc}
			\toprule
			Method & Kmeans & Otsu& Region growth & Ours\\
			\midrule
	FPS  & 0.34 & 5.47 & 3.69 & 10.85 \\
        Accuracy  & 49.50\% & 36.45\% & 44.96\% & 98.54\% \\
             \specialrule{0.08em}{1pt}{1pt}
		\end{tabular}
		\vspace{-2em}
		}
	\end{center}
\end{table}

\subsection{The Quantitative Analysis of Sky-view Images Segmentation}
Considering the high requirement of real-time and precision for vehicle positioning, we conducted quantitative tests on the efficiency and accuracy of different segmentation algorithms. The results are shown in Table \ref{tab: 2}. The efficiency of these algorithms is reflected by FPS (Frames Per Second), that FPS refers to the number of images processed per second. The Accuracy of the algorithm is reflected by “Accuracy”, which refers to the percentage of the number of correctly segmented images in the total number of processing results. The experiment is carried out on the training data set, which is convenient for us to calculate the performance index of these algorithms in sky-view images segmentation.

It can be seen from Table \ref{tab: 2} that the FCN-based image segmentation algorithm is more efficient, which is due to FCN supporting GPU acceleration. The accuracy of the other CPU (Central Processing Unit)-based machine learning methods are less than 50\%. In addition, the update cycle of the GNSS, INS and vision tightly coupled model based on MSCKF is 1s (synchronizing with the GNSS sampling frequency). Therefore, FCN's FPS fully meets the demand. Compared with machine learning methods, FCN based on deep learning is also more advantageous in terms of accuracy. Therefore, in terms of efficiency or accuracy, the FCN-based sky-view images segmentation algorithm proposed in this paper is meaningful.

\subsection{The Experimental Results of Positioning}
To verify the effectiveness of Sky-GVIO model on the positioning of car in the urban canyon, and evaluate the performance improvement of the two modes based on SPP-related and RTK-related enhanced by S-NDM. It should be noted that we use the RTK float solution. We conducted several experimental comparisons and compared against state-of-the-art methods.

\begin{figure}[t]
  % \vspace{-1em}
  \centering
  \vspace{-1mm}
  \includegraphics[width=0.45\textwidth]{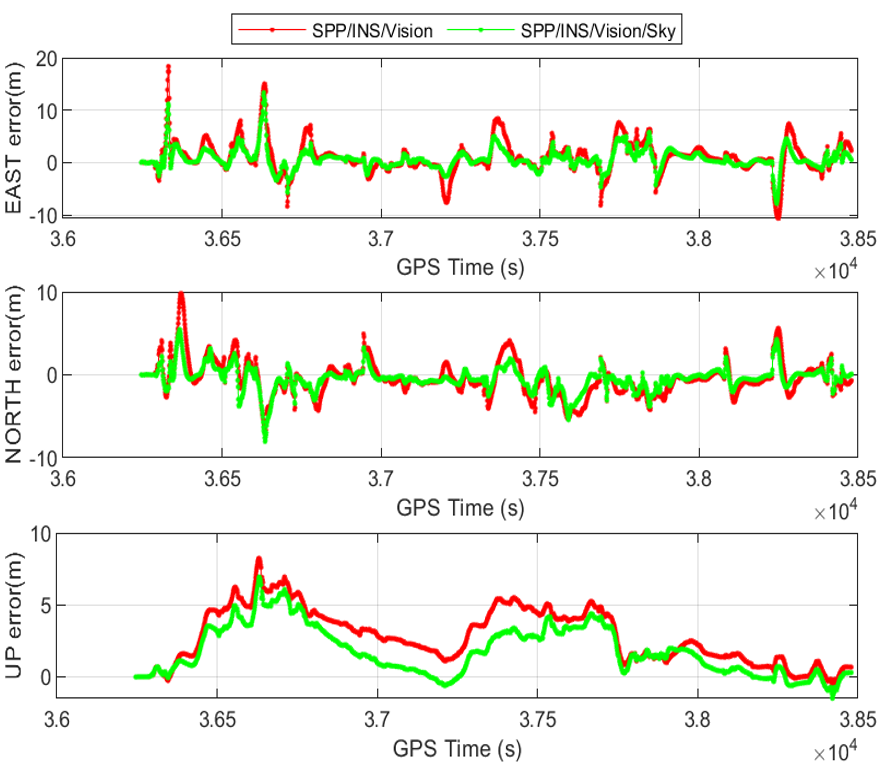}
  \vspace{-3mm}
  \caption{The comparisons of the tightly coupled (TC)-SPP/INS/Vision, SPP/INS/Vision/Sky models about the position errors in the urban canyon areas. (SPP/INS/Vision/Sky refer to Sky-GVIO of SPP-related.)}
  \label{fig:spp}
  \vspace{-4mm}
\end{figure}

\begin{figure}[t]
  % \vspace{-1em}
  \centering
  \vspace{-1mm}
  \includegraphics[width=0.45\textwidth]{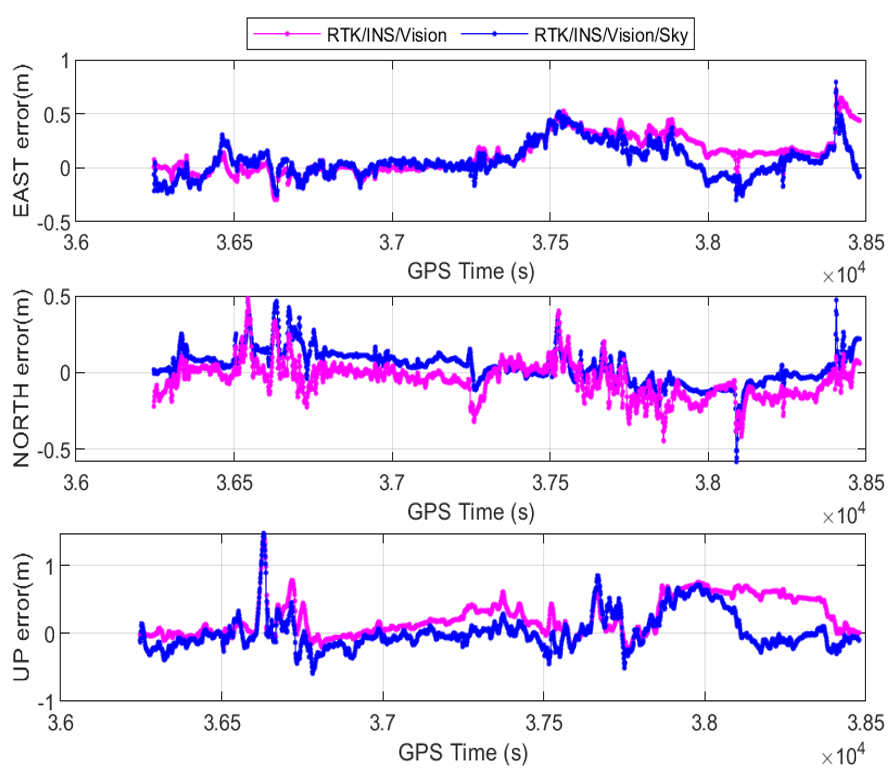}
  \vspace{-3mm}
  \caption{The comparisons of the TC-RTK/INS/Vision and TC-RTK/INS/Vision/Sky models about the position errors in the urban canyon areas. (RTK/INS/Vision/Sky refer to Sky-GVIO of RTK-related.)}
  \label{fig:rtk}
  \vspace{-4mm}
\end{figure}

\begin{table*}

\captionsetup {font={small,stretch=1.0}}
	\caption{Position RMSEs of VINS-mono, GVINS and (TC)-SPP/INS/Vision, SPP/INS/Vision/Sky, RTK/INS/Vision, RTK/INS/Vision/Sky models} 
	\label{tab: 3}
	\begin{center}
	\setlength\tabcolsep{30pt}
	\setlength{\belowcaptionskip}{1pt}
    \renewcommand\arraystretch{1.0}
	\footnotesize
 	\vspace{-2.0em}
	\resizebox{\linewidth}{!}{
\begin{tabular}{ccccc}
\cline{1-5}
&\multirow{2}{*}{Method}&\multicolumn{3}{c}{Position RMSE(m)}\\
\cline{3-5}
&&East&North&Up\\
\hline
\multirow{4}{*}{Ours}&{TC-SPP/INS/Vision}&3.24&2.14&3.39\\
&{TC-SPP/INS/Vision/Sky}&2.07&1.51&2.47\\
&{TC-RTK/INS/Vision}&0.21&0.13&0.36\\
&{TC-RTK/INS/Vision/Sky}&0.16&0.11&0.27\\
\hline
\multirow{2}{*}{Others}&{VINS-mono}&-\textit{}&-\textit{}&-\textit{}\\
&{GVINS}&2.50&1.75&2.82\\
\hline
\end{tabular}
}
\end{center}
\end{table*}

The time series of position errors for different tightly coupled models of SPP-related are presented in Fig. \ref{fig:spp} and the corresponding RMSEs are summarized in Table \ref{tab: 3}. The positioning accuracy of TC-SPP/INS/Vision in E-N-U directions is 3.24, 2.14 and 3.39 m. Different from TC-SPP/INS/Vision, TC-SPP/INS/Vision/Sky identify LOS/NLOS satellites under the GNSS challenge environment and model them to inhibit the impact of NLOS on GNSS observations. As expected, the positioning accuracy is improved to 2.07, 1.51 and 2.47 m in E-N-U directions when TC-SPP/INS/Vision enhanced by S-NDM. Compared with TC-SPP/INS/Vision, the positioning accuracy of TC-SPP/INS/Vision/Sky is improved by 36\%, 29\% and 27\% in E-N-U directions, respectively. As seen from results, TC-SPP/INS/Vision/Sky can maintain meter-level positioning accuracy. Therefore the Sky-GVIO of SPP-related is more suitable for mobile phone navigation and pedestrian navigation in urban canyons.

The time series of position errors for different tightly coupled models of RTK-related are presented in Fig. \ref{fig:rtk} and the corresponding RMSEs are summarized in Table \ref{tab: 3}. The positioning accuracy of TC-RTK/INS/Vision in E-N-U directions are 0.21, 0.13 and 0.36 m. It can be seen that the positioning accuracy of TC-RTK/INS/Vision/Sky is 0.16, 0.11 and 0.27 m in E-N-U directions which outperforms TC-RTK/INS/Vision. Compared with TC-RTK/INS/Vision, the positioning accuracy of TC-RTK/INS/Vision/Sky is improved by 24\%, 15\% and 25\% in E-N-U directions, respectively. These considerable improvements in the positioning accuracy mainly stem from GNSS NLOS detection and mitigation enhanced by sky-view images which makes the weighting of GNSS observations more reasonable.

\begin{figure}[t]
  % \vspace{-1em}
  \centering
  \vspace{-1mm}
  \includegraphics[width=0.46\textwidth]{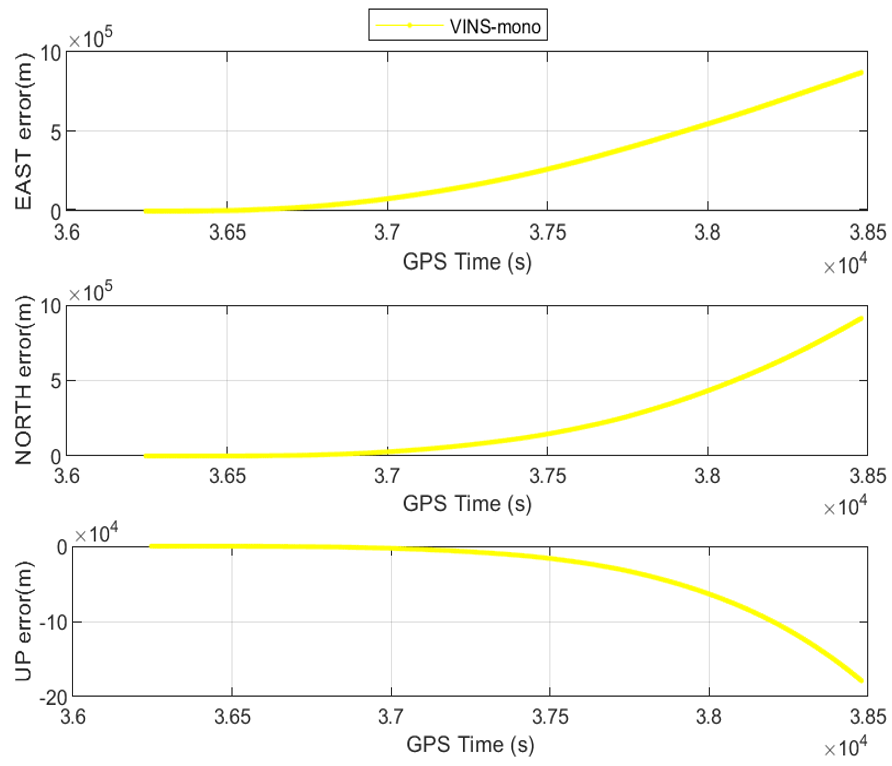}
  \vspace{-3mm}
  \caption{The position errors of VINS-mono in E-N-U directions.}
  \label{fig:vins}
  \vspace{-4mm}
\end{figure}

\begin{figure}[t]
  % \vspace{-1em}
  \centering
  \vspace{-1mm}
  \includegraphics[width=0.46\textwidth]{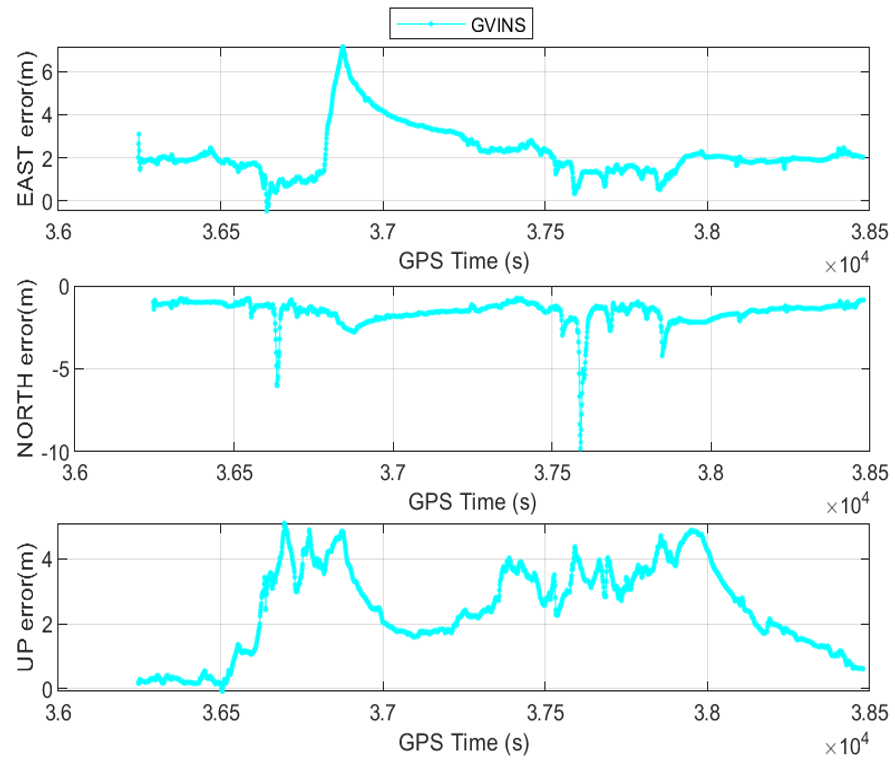}
  \vspace{-3mm}
  \caption{The position errors of GVINS in E-N-U directions.}
  \label{fig:gvins}
  \vspace{-4mm}
\end{figure}

In addition, we compared against state-of-the-art methods, including VINS-mono [5]and GVINS [7]. As we all know, VINS-mono is a very famous tightly coupled model. However, without external information for correction, VINS-mono will accumulate drift errors, resulting in gradually larger errors in E-N-U directions as shown in Fig. \ref{fig:vins}. Due to large errors obtained by VINS-mono, statistics were not carried out in Table \ref{tab: 3}. GVINS which GNSS pseudorange measurement, GNSS doppler measurement, visual constraints and inertial constraints were jointly optimized is also mature tightly coupled GNSS/INS/Vision model, which is often used to compare models of the same type. The time series of position errors for GVINS are presented in Fig. \ref{fig:gvins} and the corresponding RMSEs are summarized in Table \ref{tab: 3}. The positioning accuracy of GVINS in E-N-U directions is 2.50, 1.75 and 2.82 m which outperforms TC-SPP/INS/Vision.in this paper. This is because GVINS adds doppler measurements. However, GVINS did not carry out strict quality control in GNSS preprocessing, especially in GNSS NLOS part. Therefore, TC-SPP/INS/Vision/Sky model proposed in this paper has higher accuracy than GVINS.

\section{Conclusion} \label{sec:4}
This paper presents a GNSS NLOS-detectable, reliable tight-coupled model in urban canyons, called Sky-GVIO.We detail a module for GNSS NLOS detection and mitigation, and extend it to the tightly coupled GNSS/INS/Vision model. Based on this, we evaluate the position performance of the SPP-related and RTK-related tight-coupled models. We find that these models can be helped to improve the positioning accuracy by the S-NDM algorithm proposed in this paper.In urban canyon environments where GNSS performance is challenging, our Sky-GVIO model of RTK-related can achieve sub-decimeter accuracy, which is exciting for users with high-precision location service needs. In addition, our Sky-GVIO model of SPP-related also achieves meter-level positioning accuracy in this GNSS-challenged urban canyon environment, which is also very meaningful for low-cost users such as cell phone navigation and pedestrian navigation.

In the future, we still have the following work to do:

(1) Enhancing the utilization of fish-eye camera data beyond GNSS NLOS detection, potentially integrating fish-eye camera observations into the proposed model.

(2) Accuracy is expected to reach centimeter level. By adding prior information (such as high-precision maps), the whole system is more robust and the positioning accuracy is higher.

\section*{Acknowledgments}
Upon reasonable request to the corresponding author, the experimental data used in this research is available. 
This work was supported by the National Key Research and Development Program of China under Grant 2021YFB2501100.

% \newpage

% {\appendix[Proof of the Zonklar Equations]
% Use $\backslash${\tt{appendix}} if you have a single appendix:
% Do not use $\backslash${\tt{section}} anymore after $\backslash${\tt{appendix}}, only $\backslash${\tt{section*}}.
% If you have multiple appendixes use $\backslash${\tt{appendices}} then use $\backslash${\tt{section}} to start each appendix.
% You must declare a $\backslash${\tt{section}} before using any $\backslash${\tt{subsection}} or using $\backslash${\tt{label}} ($\backslash${\tt{appendices}} by itself
%  starts a section numbered zero.)}


% \section{References Section}
% You can use a bibliography generated by BibTeX as a .bbl file.
%  BibTeX documentation can be easily obtained at:
%  http://mirror.ctan.org/biblio/bibtex/contrib/doc/
%  The IEEEtran BibTeX style support page is:
%  http://www.michaelshell.org/tex/ieeetran/bibtex/
 

% You can manually copy in the resultant .bbl file and set second argument of $\backslash${\tt{begin}} to the number of references
%  (used to reserve space for the reference number labels box).

% \bibliographystyle{IEEEtran}
% \end{thebibliography}
\bibliographystyle{IEEEtran}
\bibliography{ref}

% \newpage

 

% \newpage

\begin{IEEEbiography}[{\includegraphics[width=1in,height=1.25in,clip,keepaspectratio]{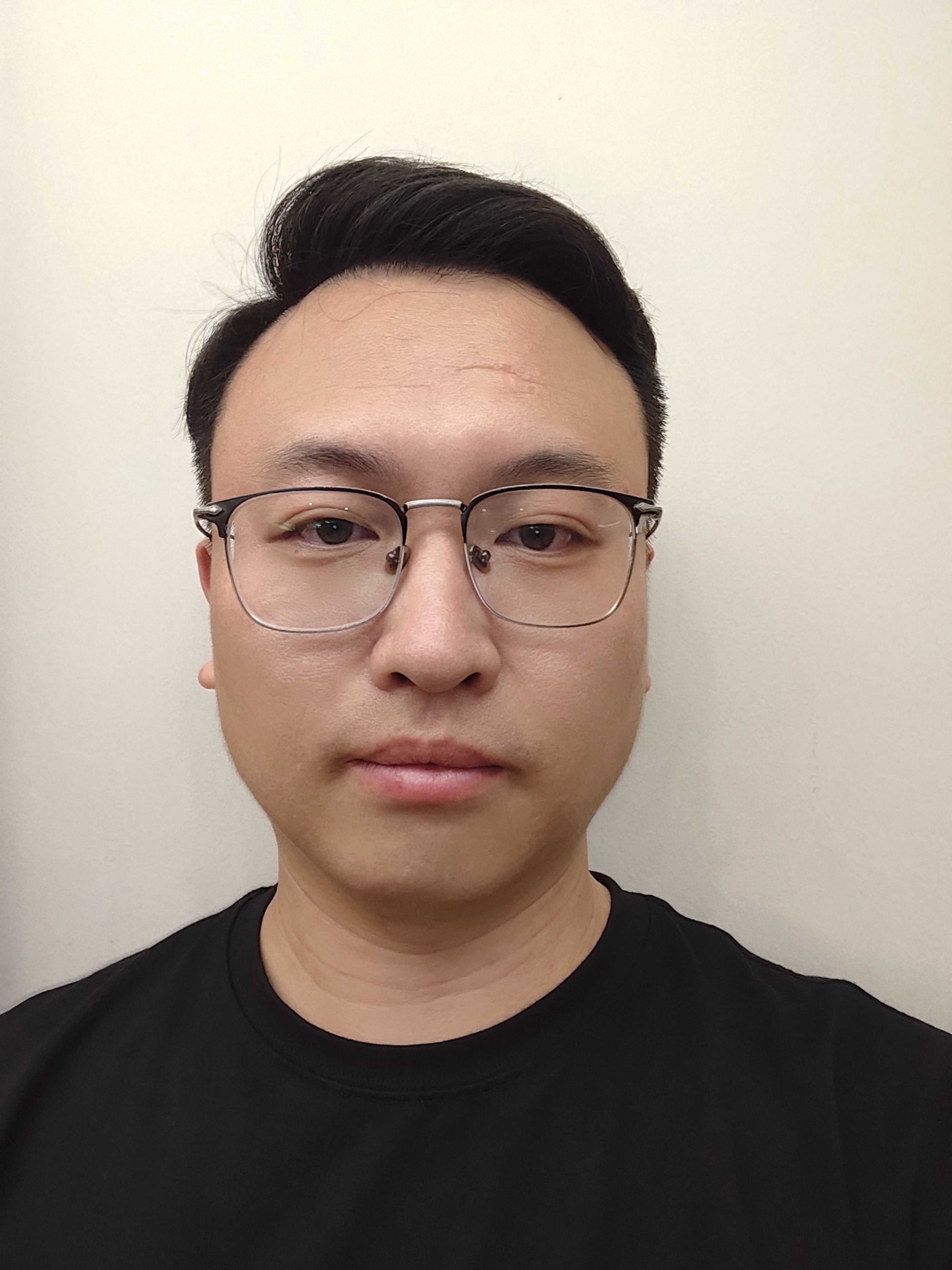}}]{Jingrong Wang}  received the B.E. degree in the School of Land Science and Technology from China University of Geosciences, Beijing, China, in 2014 and 2018, and the M.E. degree in  the State Key Laboratory of Information Engineering in Surveying, Mapping and Remote Sensing (LIESMARS) from Wuhan University, Wuhan, China, in 2018 and 2020, respectively, where, he is currently pursuing the Ph.D degree with the GNSS Research Center. His research interests include GNSS precise positioning, visual inertial odometry (VIO) and multi-sensor fusion algorithm. 
\end{IEEEbiography}

\begin{IEEEbiography}[{\includegraphics[width=1in,height=1.25in,clip,keepaspectratio]{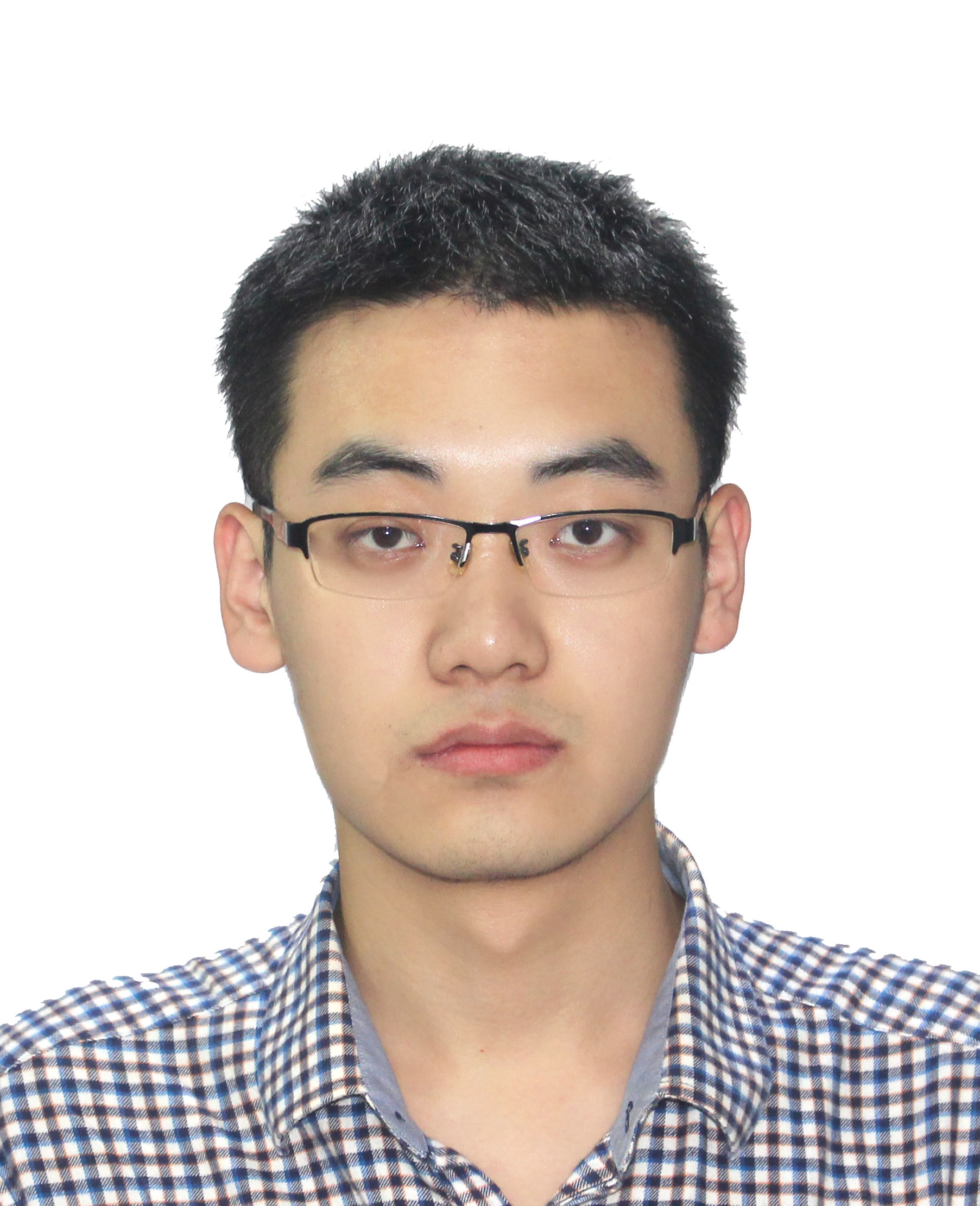}}]{Bo Xu} received the B.E. degree in the School of Land Science and Technology from China University of Geosciences, Beijing, China, in 2014 and 2018, and the M.E. degree in the School of Geodesy and Geomatics from Wuhan University, Wuhan, China, in 2018 and 2021, respectively, where, he is currently pursuing the Ph.D degree with the School of Geodesy and Geomatics. His research interests include GNSS precise positioning, visual SLAM, visual inertial odometry (VIO) and multi-sensor fusion algorithm.
\end{IEEEbiography}

\begin{IEEEbiography}[{\includegraphics[width=1in,height=1.25in,clip,keepaspectratio]{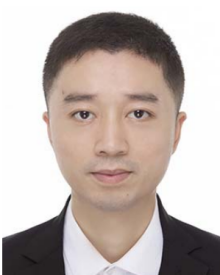}}]{Ronghe Jin}  received the B.S. degree in geographic information system and the M.S. degree in surveying and mapping from Wuhan University, Wuhan, China, in 2013 and 2016, respectively. He received Ph.D. degree with the School of Geodesy and Geomatics, from Wuhan University, Wuhan, China, in 2022. He is currently a Postdoctoral Fellow with the Hong Kong Polytechnic University, Hong Kong, China. His research interests include GNSS precise positioning, visual SLAM, visual inertial odometery (VIO), multi-sensor fusion algorithm, and their applications in land vehicle navigation. 
\end{IEEEbiography}

\begin{IEEEbiography}[{\includegraphics[width=1in,height=1.25in,clip,keepaspectratio]{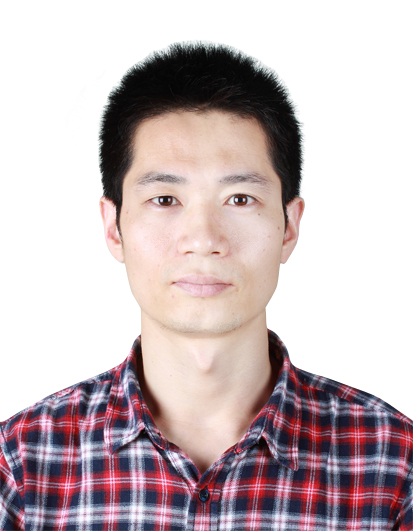}}]{Shoujian Zhang} received the B.Eng. and Ph.D.
degrees in the School of Geodesy and Geomatics from Wuhan University, Wuhan, China, in 2004 and 2009, respectively. He is an associate professor at the school of geodesy and geomatics from Wuhan University. His research interests are GNSS data processing,  multiple sensor  data fusion algorithms and applications.

\end{IEEEbiography}

\begin{IEEEbiography}[{\includegraphics[width=1in,height=1.25in,clip,keepaspectratio]{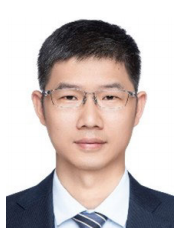}}]{Xingxing Li} received the B.Eng. degree in geodesy
and survey engineering from the School of Geodesy and Geomatics, Wuhan University, Wuhan, China, in 2008, and the Ph.D. degree in geodesy from the Department of Geodesy and Remote Sensing, German Research Center for Geosciences (GFZ),
Potsdam, Germany, in 2015. He is currently a Professor with Wuhan University.
His current research interests include GNSS precise data processing and multisensor fusion navigation.
\end{IEEEbiography}

\begin{IEEEbiography}[{\includegraphics[width=1in,height=1.25in,clip,keepaspectratio]{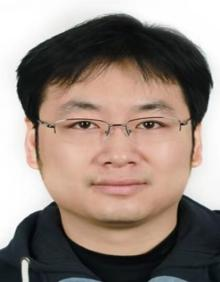}}]{Kefu Gao} received his Ph.D. degree in informaiton security from Wuhan University in 2016. From October 2011, he worked at GNSS Research Center of Wuhan University. His research interests include the theories and methods of location based service, intelligent computing.
\end{IEEEbiography}

\begin{IEEEbiography}[{\includegraphics[width=1in,height=1.25in,clip,keepaspectratio]{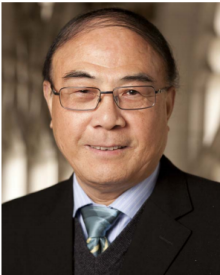}}]{Jingnan Liu} received the B.Eng. degree in astronomical geodesy and the M.Eng. degree in geodesy from the Wuhan Technical University of Surveying and Mapping, China (now Wuhan University) in 1967 and 1982, respectively. He is an Academician of the Chinese Academy of Engineering, a Former President of Wuhan University, the first Chancellor of Duke Kunshan University, and the Director of the National Engineering Research Center for Satellite Positioning System. He has been engaged in the research of geodetic theories and applications, including national coordinate system establishment, global navigation satellite system (GNSS) technology, and software development as well as large project implementation. As an academic authority in the GNSS field, he has been awarded more than ten national or provincial prizes for progress in science and technology.
\end{IEEEbiography}

\vfill

\end{document}